\pdfoutput=1

\documentclass[11pt]{article}

\usepackage{acl}

\usepackage{times}
\usepackage{latexsym}

\usepackage[T1]{fontenc}

\usepackage[utf8]{inputenc}

\usepackage{microtype}

\usepackage{graphicx}
\usepackage{subfigure}
\usepackage{amsmath}
\usepackage{multirow}
\usepackage{makecell}
\usepackage{booktabs}

\title{SelfMix: Robust Learning Against Textual Label Noise with\\ Self-Mixup Training}

\author{Dan Qiao$^{1}$, Chenchen Dai$^1$, Yuyang Ding$^1$, Juntao Li$^1$,\\ 
\textbf{Qiang Chen$^2$, Wenliang Chen$^{1}$, Min Zhang$^{1}$}\\
 $^1$Institute of Computer Science and Technology, Soochow University, China\\
 $^2$Alibaba Group\\
 \texttt{\{danqiao.jordan,morningcc125,yyding.me\}@gmail.com}\\
\texttt{\{ljt,wlchen,minzhang\}@suda.edu.cn};  \texttt{lapu.cq@alibaba-inc.com}
  }

\begin{document}
\maketitle
\begin{abstract}
The conventional success of textual classification relies on annotated data, and the new paradigm of pre-trained language models (PLMs) still requires a few labeled data for downstream tasks.
However, in real-world applications, label noise inevitably exists in training data, damaging the effectiveness, robustness, and generalization of the models constructed on such data.
Recently, remarkable achievements have been made to mitigate this dilemma in visual data, while only a few explore textual data.
To fill this gap, we present SelfMix, a simple yet effective method, to handle label noise in text classification tasks.
SelfMix uses the Gaussian Mixture Model to separate samples and leverages semi-supervised learning.
Unlike previous works requiring multiple models, our method utilizes the dropout mechanism on a single model to reduce the confirmation bias in self-training and introduces a textual level mixup training strategy.
Experimental results on three text classification benchmarks with different types of text show that the performance of our proposed method outperforms these strong baselines designed for both textual and visual data under different noise ratios and noise types.
Our code is available at \url{https://github.com/noise-learning/SelfMix}. 
\end{abstract}

\section{Introduction}

The excellent performance of deep neural networks~(DNNs) depends on data with high-quality annotations. 
However, data obtained from the real world is inevitably mixed with wrong labels~\citep{data1,data2,data3}.
Models trained on these noisy datasets would easily overfit the noisy labels~\citep{noise-effect, ELR}, especially for pre-trained large models~\cite{zhang2021commentary}, and the performance will be negatively affected.

Research on learning with noisy labels (LNL) has gained popularity. 
Previous work has revealed that clean samples and noisy samples play different roles in the training process and behave differently in terms of loss values or convergence speeds etc.~\citep{ELR}.
Different types of noise have different effects on the training. 
For instance, the impact of class-conditional noise (CCN) can simulate the confusion between similar classes, and the effect of instance-dependent noise (IDN) can be more complex.

Most of the current methods perform experiments on visual data. 
Label noise on visual data often goes against objective facts and is easy to distinguish. 
As for NLP, there may be disagreement even among expert annotators due to the complexity of semantic features and the subjectivity of language understanding. 
For example, suppose there is a piece of news about ``The Economic Benefit of Competitive Sports to our Cities''. 
In that case, it is hard to tell whether it belongs to Economic news or Sports news without fully understanding the contextual information.
Although a few works pay attention to the natural language area, their methods are mostly based on the trained-from-scratch models like LSTM and Text-CNN~\citep{CIKM21,NAACL2019}. 
However, PLMs might be a better choice since the whole training process can be divided into two stages, and the wrong labels do not corrupt the pre-training process.
Table~\ref{tab:basemodel} makes comparisons between PLMs and traditional networks on the robustness against label noise.

In conclusion, it is vital to explore how to learn with noisy labels on textual data and use the robust PLMs as the base model.
This paper proposes SelfMix, i.e., a self-distillation robust training method based on the pre-trained models.
Section~\ref{sec:2} introduces some related works and explains the motivation of our proposed method.

Our contributions can be concluded as follows:
\begin{itemize}
    \item
    We propose SelfMix, a simple yet effective method to help learn with noisy labels, which utilizes a self-training approach. 
    Our method only needs a single model and utilizes a mixup training strategy based on the aggregated representation from pre-trained models.
    \item
    We perform comprehensive experiments on three different types of text classification benchmarks under various noise settings, including the challenging instance-dependent noise, which is usually ignored in other works on textual data, which demonstrate the superiority of our proposed method over strong baselines.
\end{itemize}

\section{Related Work}
\label{sec:2}
\textbf{Learning with Noisy Labels.}
A direct yet effective idea to handle label noise is to find the noisy samples and reduce their influence by resampling or reweighting~\citep{DNNitself1}. 
\citet{mentornet} train another neural network to provide a curriculum to help StudentNet focus on the samples whose labels is probably correct.
\citet{coteaching} jointly train two deep neural networks and feed each model the top $r$\% samples with the lowest loss evaluated by the other model in each mini-batch. 
Following~\citet{coteaching}, \citet{coteaching+} explore how disagreement can help the model.
Some researchers believe that there exists a transition from ground-truth label distribution to the noisy label distribution and estimate the noise transition matrix to absorb this transition~\citep{adaptation-ntm}.
\citet{confidentlearning} directly estimate the joint distribution matrix between the noisy labels and real labels.
\citet{CIKM21} use a fully connected layer to capture the distribution transition.
However, most of these methods either need model ensembling or require cross-validation, which is time-consuming and needs multiple parameters.

Some other works focus on designing a more robust training strategy.
Since DNNs with Cross-Entropy loss tend to overfit noisy labels~\citep{CE_cons1}, some researchers redesign noise-robust loss functions~\citep{SCE, GCE, MAE, DMI}.
When trained on noisy data, DNNs tend to learn from the clean data during an ``early learning'' phase before eventually memorizing the wrong data~\citep{earlylearning1, earlylearning3}, based on which~\citet{ELR} offer an easy regularization capitalizing on early learning.
Some other works like~\citet{CDR} find that only partial parameters are essential for generalization, which offers us a new perspective to reconsider what difference exactly the noisy labels make to the model's learning.
This kind of approach treats all samples indiscriminately thus the performance is sometimes unsatisfactory under a high noise ratio.

Some excellent work combines these two ideas ~\citep{semisup1, divmix}.
\citet{CIKM21} add an auxiliary noise model $N_M$ over the classifier to predict noisy labels and jointly train the classifier and the noise model through a de-noising loss function.
\citet{CORE} progressively sieve out corrupted examples and then leverage semi-supervised learning.

\textbf{Mixup Training.}
Mixup training ~\citep{mixup} is a widely used data-augmentation method to alleviate memorization and sensitivity to adversarial samples on visual data. 
It combines the inputs and targets of two random training samples to generate augmented samples.
However, applying mixup on textual data is a great challenge since linear interpolations on discrete inputs damage the semantic structure.
Some literature has explored the textual mixup mechanism like:~\citet{tmix} propose to mix the hidden vector in the last few encoder layers;~\citet{ssmix} find a new way to combine two texts which can also be treated as a data augmentation strategy.
In this paper, we do not make comparisons for the following reasons: (1) Our EmbMix is simpler in practical use and there is little difference in the final performance of various methods according to~\citet{tmix}. (2) Some other methods need data augmentation while EmbMix does not.

\textbf{Proposed Method.}
Since simply redesigning a robust loss function tends to have poor performance under a high noise ratio, we combine sample selection with the robust training methods.
Unlike the previous work that needs model ensembling or uses cross-validation, we train a single network with dropout to reduce confirmation bias in self-training.
We make following improvements regarding to the characteristics of the textual data: (1) The decision boundaries in image-classification tasks are more clear.
However, the main idea of the same text can vary under different contexts and sometimes there is even no absolute correct label. So we iteratively use the Gaussian Mixture Model (GMM) to fit the loss distribution and use the predicted soft label to replace the label of the fusing data rather than setting a threshold and arbitrarily discarding the undesired samples at the beginning.
(2) Unlike the pixel input of visual data, the input of text is discrete. 
So for the separated data, we leverage a manifold mixup training strategy based on the aggregated representation from the PLMs.

\section{Methodology}
\begin{figure}[ht]
\includegraphics[width=1.0\columnwidth]{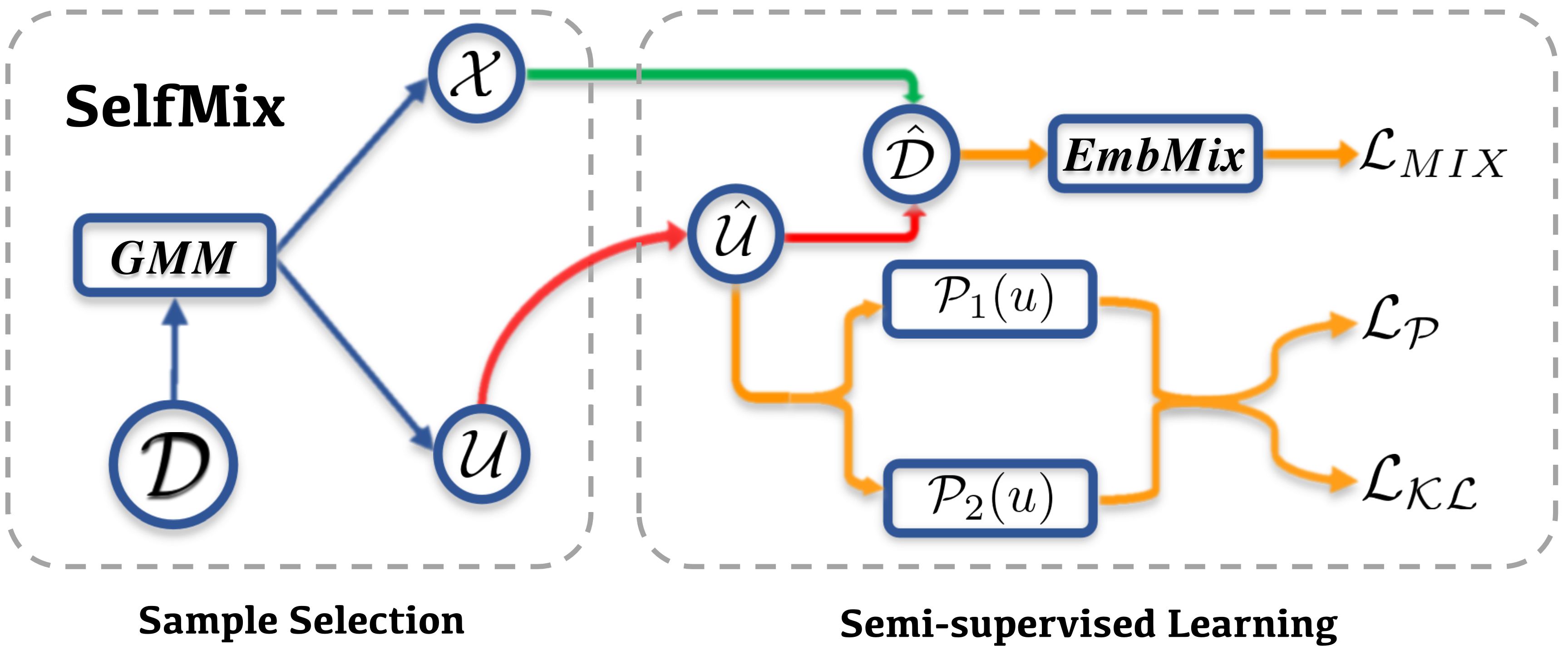}
\caption{The overall framework of SelfMix}
\label{fig:figure1}
\end{figure}
\noindent Figure~\ref{fig:figure1} shows an overview of our proposed SelfMix.
Our method first uses GMM to select the samples that are more likely to be wrong and erase their original labels.
Then we leverage semi-supervised learning to jointly train the labeled set $\mathcal{X}$~(contains mostly clean samples) and an unlabeled set $\mathcal{U}$~(contains mostly noisy samples).
We also introduce a manifold mixup strategy based on the hidden representation of the [CLS] token named EmbMix. 

\subsection{Preliminary}
In real-world data collection, the observed labels are often corrupted.
So the only difference between this task and the traditional text classification task is that a certain proportion of incorrect labels exist in training samples.
Let $\mathcal{D}=\left\{(x_i,y_i)\right\}^N_{i=1}$ denote the original dataset, where $N$ is the number of samples, $x_i$ is the text of the $i^{th}$ sample, and $y_i$ is the one-hot representation of the observed label of the $i^{th}$ sample. 
For the base model, we denote $\theta$ as the parameters of the pre-trained encoder model and $\phi$ as the parameters of MLP classifier head with 2 fully connected layers. 
The standard optimization method tries to minimize the empirical risk by applying the cross-entropy loss:
\begin{equation}
\centering
\hspace{-0.04cm}
\mathcal{L}=\left\{ \ell_{i} \right\}_{i=1}^N=\left\{ - y_{i}^{T} \log \left(p\left(x_{i};\theta,\phi\right)\right) \right\}_{i=1}^N,
\end{equation}
where $p\left(x;\theta\right)$ denotes the softmax probability of the model output.
We first warm up the model using $\mathcal{L}$ to make it capable of doing preliminary classification tasks without overfitting noisy labels and then perform SelfMix for the rest epochs.

\subsection{Sample Selection}
On noisy data, Deep neural networks will preferentially learn simple and logical samples first and reduce their loss. 
Namely, noisy samples tend to have a higher loss in the early stage ~\citep{earlylearning3}. 
Preliminary experiments show that the loss distributions of clean and noisy samples during training tend to subject to two Gaussian Distributions, where the loss of the clean samples hold a smaller mean value.
Taking advantage of such training phenomena, we apply the popular used Gaussian Mixture Model~\citep{bmmgmm} to distinguish noisy samples by feeding the per-sample loss.
For IDN, noisy labels rely on both input features and underlying true labels, so the noise in each class is different, making the loss scales from different classes vary greatly.
The relatively high-loss samples in low-loss class may also be treated as clean samples.
So we compute a class-regularization loss instead of the standard cross-entropy loss, which can better model the distributions in IDN.
For each class $c$, the set $\mathcal{L}_c=\left\{\ell_i \mid y_i=c, i\in[N] \right\}$ contains the cross-entropy loss values of all samples with label $c$, then $\mu_c$ and $\sigma_c$ denote the arithmetic mean and standard deviation of $\mathcal{L}_c$ respectively. Our regularization loss has the following form:
\begin{equation}
\mathcal{L}'=\left\{\ell'_i\right\}^N_{i=1}=\left\{(\ell_i-\mu_{y_i})/\sigma_{y_i}\right\}^N_{i=1}.
\end{equation}
We feed the loss $\mathcal{L}$~($\mathcal{L}'$ for IDN) to a 2-component GMM and use Expectation-Maximization (EM) algorithms to fit the GMM to the observations.
Let $w_i=p(g|\ell'_{i})$ represent the probability of the $i^{th}$ sample belonging to the Gaussian component with smaller mean $g$, which can also be considered as the clean probability due to the small-loss theory~\citep{earlylearning1}.
By setting the threshold $\tau$ for the probability $w_i$, we can divide the original dataset $\mathcal{D}$ into a labeled set $\mathcal{X}$ and an unlabeled set $\mathcal{U}$ where the labels of samples that are more likely to be wrong will be erased:
\begin{equation}
\begin{aligned}
\mathcal{X}&=\left\{\left(x_{i}, y_{i}\right) \mid x_{i} \in \mathcal{D}, w_{i} \geq \tau\right\}, \\
\mathcal{U}&=\left\{\left(x_{i} \right) \mid x_{i} \in \mathcal{D}, w_{i}<\tau\right\}.
\end{aligned}
\end{equation}

\subsection{Semi-supervised Self-training}
To make semi-supervised learning work better, we first do pre-process on the unlabeled set.
For the unlabeled set $\mathcal{U}$, the original label is most likely wrong and has been discarded. 
Therefore, we generate the soft label $\hat{y}$ by sharpening the model's predicted distribution, making the distribution more concentrated ~\citep{self-training}.
\begin{align}
\hat{y}&=\text{Sharpen}\left(p\left(x;\theta,\phi\right)\right), \\
\hat{\mathcal{U}}&=\left\{\left(x_{i}, \hat{y}_{i}\right) \mid x_{i} \in \mathcal{U}\right\}, \\
\hat{\mathcal{D}}&=\mathcal{X} \cup \hat{\mathcal{U}}.
\end{align}
Here $\text{Sharpen}\left(\cdot\right)$ is the temperature sharpening function commonly used in self-training. 
$\hat{\mathcal{D}}$ contains the clean samples with original labels and noisy samples with predicted soft labels.

\textbf{Textual Mixup based on EmbMix.}
\ \ Mixup training strategy is widely used in semi-supervised learning and noise-robust training ~\citep{mixup}. 
It applies linear interpolation to the input vectors and associated targets.
Although image data can be mixed on the pixel level, mixing the discrete word index makes no sense for text classification.
Considering that the {\tt [CLS]} embedding encoded by PLMs has the ability of semantic representations, we attempt to apply interpolations on the {\tt [CLS]} embedding.
Specifically, randomly choose two samples $(x_i,y_i)$, $(x_j,y_j)$ and the mixed sample $(e^{\prime}_i,y^{\prime}_i)$ is defined as:
\begin{align}
\lambda  &\sim \operatorname{Beta}(\alpha, \alpha), \label{eq:m1} \\ 
\lambda^{\prime} &=\max (\lambda, 1-\lambda), \\
e_k & = \text{Encoder}\left(x_k;\theta\right), \\
e^{\prime}_i &=\lambda^{\prime} e_{i}+\left(1-\lambda^{\prime}\right) e_{j}, \\
y^{\prime}_i &=\lambda^{\prime} y_{i}+\left(1-\lambda^{\prime}\right) y_{j}, \label{eq:m2}
\end{align} 
where $\text{Encoder}\left(x;\theta\right)$ denotes the sentence {\tt [CLS]} embeddings obtained by pre-trained models. 

Finally, the EmbMix method for dataset $\hat{\mathcal{D}}$ is as follows:
\begin{equation}
\begin{aligned}
\tilde{\mathcal{D}}=\left\{\left(e^{\prime}_i,y^{\prime}_i\right) \mid \left(x_i,y_i\right),\left(x_j,y_j\right) \in \hat{\mathcal{D}} \right\},
\end{aligned}
\end{equation}
where $(e^{\prime}_i,y^{\prime}_i)$ is computed by eq.(\ref{eq:m1}-\ref{eq:m2}).

\subsection{Loss Function}
\textbf{Mix-Loss.}
Given our augmented dataset $\tilde{\mathcal{D}}$ obtained by EmbMix, we use the standard cross-entropy loss for semi-supervised learning:
\begin{align}
\mathcal{L}_{MIX}&=-\frac{1}{\mid \tilde{\mathcal{D}} \mid} \sum_{(e,y)\in \tilde{\mathcal{D}}} y^T \log \left(p\left(e;\phi\right)\right).
\end{align}
Here $p(e;\phi)$ denotes the predicted probability of the mixed target using the mixed hidden representation $e$ as the input.

\textbf{Pseudo-Loss.} According to the Low-density Separation Assumption theory, the decision boundary of a classifier should preferably pass through low-density regions in the input space ~\citep{entropy-theory}. 
To achieve this, we add a special regularization on the unlabeled set to penalize those samples whose output probability value of the predicted class is small:
\begin{align}
\tilde{y}_i&=\arg \max (p\left(x_{i};\theta,\phi)\right),\\
\mathcal{L}_\mathcal{P}&=-\frac{1}{\mid {\mathcal{U}} \mid}\sum_{x_i \in {U}}\tilde{y}_i log(p\left(x_{i};\theta,\phi)\right).
\end{align}
Here $p(x_{i};\theta,\phi)$ denotes the model's prediction of sample $x_i$, and $\tilde{y}$ denotes the one-hot representation of the pseudo-label that the model predicts.
Preliminary experiments show that pseudo-loss regularization is more effective than a simple entropy-minimization. 

\textbf{Self-consistency Regularization.} 
It is worth mentioning that confirmation bias caused by error accumulation is common in self-training. 
Model ensembling is a widely used method to handle this.
Dropout~\citep{dropout} mechanism can be seen as an implicit sub-models ensembling.
So we use dropout when training the network and close dropout when making sample selection or inference.
Label noise under a high noise ratio setting blurs the decision boundaries between classes, leading to a severe inconsistency between sub-models. 
So we add R-Drop~\citep{rdrop} loss, a simple but effective dropout regularization method to constrain the consistency of these sub-models:
\begin{equation}
\begin{aligned}
\mathcal{L}_{R} & =\sum_{x \in \mathcal{U}}\frac{1}{2}( 
\mathcal{D}_{KL}\left( p_1\left(x;\theta,\phi \right) ||~p_2\left(x;\theta,\phi\right) \right)  
\\ & + \mathcal{D}_{KL}\left( p_2\left(x;\theta,\phi \right) ||~p_1\left(x;\theta,\phi\right) \right)),
\end{aligned}     
\end{equation}
where $p_1(x;\theta,\phi)$ and $p_2(x;\theta,\phi)$ are two predicted distributions obtained by feeding the same sample twice, $\mathcal{D}_{KL}\left(a || b \right)$ computes the Kullback-Leibler divergence between two probability distributions.

Finally, the total loss for SelfMix is:
\begin{equation}
\mathcal{L} = \mathcal{L}_{MIX}+\lambda_p \mathcal{L}_\mathcal{P}+\lambda_r \mathcal{L}_{\mathcal{R}},
\end{equation}
where $\lambda_p$ and $\lambda_r$ are the hyper-parameters to control the weight of the extra loss.

\section{Experiments}
\label{sec:expirement}

\subsection{Settings}
\textbf{Datasets and Noise Settings.} 
We do experiments on three text classification benchmarks of different types, including Trec~\citep{Trec}, AG-News~\citep{agnews}, and IMDB~\citep{IMDB}~(Table~\ref{table:data}). 
In the preliminary experiments, we find that PLMs are robust to random noise on textual data (Table ~\ref{tab:basemodel}).
The test accuracy drops by only 3\% even under 40\% random noise, which may benefit from the powerful pre-trained knowledge.
So we evaluate our strategy under the following two types of label noise:
\begin{itemize}
\item Asymmetric noise (Asym): Asymmetric noise tries to simulate the mislabeling between classes. For a given class, we follow ~\citet{INCV} and choose a certain proportion of samples and flip their labels to the corresponding class according to the asymmetric noise transition matrix.

\item Instance-dependent noise (IDN): The probability of being mislabeled depends on the feature of instances. So we use the other trained model as the feature extractor.
The labels of the samples that are closest to decision boundaries are flipped to their counter class as noisy labels ~\citep{noise-effect}, which is more challenging and quite realistic.
\end{itemize}

\begin{table}\small
\centering
\renewcommand{\arraystretch}{1.1}
\setlength\tabcolsep{4pt}
  \begin{tabular}{lcccc}
    \toprule
    \textbf{Name}& \textbf{Class}& \textbf{Type}& \textbf{Train} & \textbf{Test} \\
    \midrule
    Trec& 6 & Question-Type & 5452 & 500\\
    IMDB & 2 & Sentiment Analysis& 45K & 5K\\
    AG-News & 4 & News Categorization& 120K & 7.6K\\
  \bottomrule
\end{tabular}
  \caption{The statistics of datasets.}
\label{table:data}
\end{table}

\begin{table}\small
\centering
\renewcommand{\arraystretch}{1.1}
\setlength\tabcolsep{3.5pt}
\begin{tabular}{l|ccc|ccc}
\hline
\textbf{Dataset} &\multicolumn{3}{c|}{\textbf{Trec}}& \multicolumn{3}{c}{\textbf{AG-News}} \\
\hline
\hline
Rand (\%) & 0 & 20 & 40 & 0 & 20 & 40 \\
\hline
BERT        & 97.04 & 95.75 & 94.07 & 94.03 & 93.19 & 92.51 \\
\hline
\hline
Asym (\%) & 0 & 20 & 40 & 0 & 20 & 40 \\
\hline
Text-CNN    & 93.48 & 88.36 & 70.52 & 90.83 & 88.95 & 76.69 \\
\hline
LSTM        & 92.58 & 90.68 & 83.96 & 91.92 & 90.20 & 88.62 \\
\hline
BERT        & 97.04 & 95.52 & 89.04 & 94.03 & 93.38 & 91.59 \\
\hline
RoBERTa     & 96.92 & 96.32 & 92.12 & 94.10 & 93.91 & 92.74 \\
\hline
\end{tabular}
\caption{ Preliminary experiments (\%) for different base models under symmetric and asymmetric noise.
}
\label{tab:basemodel}
\end{table}

\begin{table*}[ht]\small
\centering
\renewcommand{\arraystretch}{1.2}
  \renewcommand\tabcolsep{5pt}
  \begin{tabular}{lc|cc|cccc|cccl}
    \toprule
    Dataset~/~$(\lambda_p,\lambda_r)$&&\multicolumn{2}{c|}{Trec~$(0.2,0.3)$}&\multicolumn{4}{c|}{AG-News~$(0.2,0.3)$}&\multicolumn{4}{c}{IMDB~$(0.1,0.5)$}\\
    \midrule
    Data Size&&\multicolumn{2}{c|}{5,453~(All)} &\multicolumn{2}{c}{5,000}&\multicolumn{2}{c|}{120,000~(All)} &\multicolumn{2}{c}{5,000}&\multicolumn{2}{c}{45,000~(All)}\\
    \midrule
    Noise Ratio (\%)& &20 &40 &20 &40 &20 &40 &20 &40 &20 &40 \\
    \midrule
    \multirow{2}*{BERT} 
            & best &95.52&89.04&89.55&80.90&93.38&91.59&88.51&80.81&92.67&87.70\\ 
            & last &93.48&69.88&84.40&62.33&90.32&74.04&81.20&63.55&87.40&61.82\\
    \hline
    \multirow{2}*{BERT+Co-Teaching} 
            & best &95.96&92.76&89.70&87.24&93.43&92.03&88.81&84.39&92.94&88.45\\
            & last &95.32&90.08&88.77&82.53&93.01&85.03&88.24&82.68&91.68&84.43\\
    \hline
    \multirow{2}*{BERT+Co-Teaching+} 
            & best &\textbf{96.37}&91.14&89.45 &85.81  & 92.93&90.96& 88.57&81.75 &92.71&87.94\\
            & last &\textbf{95.98}&87.24&89.12 &79.82  & 92.87&90.41& 88.33&81.23 &92.69&87.07\\
    \hline
    \multirow{2}*{BERT+SCE} 
            & best &94.72&91.28&89.62&86.72&93.13&90.78&88.76&82.65&92.82&87.32\\
            & last &94.04&82.44&89.43&74.37&93.03&87.34&87.74&74.38&92.77&82.52\\
    \hline
    \multirow{2}*{BERT+ELR} 
            & best &96.08&92.16&89.88&85.43&\textbf{93.63}&92.00&88.70&82.45&93.13&87.62\\
            & last &95.40&88.28&89.47&81.24&93.30&90.67&87.76&72.71&92.50&79.54\\ 
    \hline
    \multirow{2}*{BERT+Confident-Learning} 
            & best &95.92&91.80&89.83&84.77&93.57&91.96&89.05&81.65&92.66&87.13\\
            & last &95.36&88.64&89.27&78.48&93.38&89.97&88.62&77.93&92.52&83.39\\
    \hline
    \multirow{2}*{BERT+NM-Net} 
            & best &96.00&90.92&89.35&81.35&93.54&92.09&88.70&81.21&92.93&88.47\\
            & last &94.84&79.76&85.41&63.26&\textbf{93.47}&84.55&88.41&74.62&92.28&86.60\\       
    \hline
    \multirow{2}*{BERT+SelfMix} 
    & best &\textbf{96.32}&\underline{\textbf{94.12}}&\textbf{89.90}&\underline{\textbf{88.80}}&93.39&\textbf{92.79}&\textbf{89.20}&\underline{\textbf{86.38}}&\textbf{93.30}&\underline{\textbf{90.19}}\\
    & last &\underline{\textbf{96.04}}&\underline{\textbf{93.80}}&\underline{\textbf{89.79}}&\underline{\textbf{88.63}}&\underline{93.04}&\underline{\textbf{92.40}}&\underline{\textbf{88.84}}&\underline{\textbf{86.38}}&\underline{\textbf{92.86}}&\underline{\textbf{90.12}}\\
  \bottomrule
\end{tabular}
  \caption{Average test accuracy~(\%) of five runs on the Trec, AG-News, and IMDB datasets with different data sizes under different ratios of asymmetric noise. The results with outstanding improvement over the base model are bolded, and underline values indicate the statistically significantly better (by paired bootstrap test, $p < 0.05$) performances than BERT.}
\label{table:asym}
\end{table*}

\textbf{Model Architectures.}
Most related works perform experiments based on trained-from-scratch models, while PLMs have been shown to have great potential for all kinds of language tasks. 
Thus we conduct experiments on different models to evaluate their robustness against label noise.
Table \ref{tab:basemodel} shows that the pre-trained model is more robust than traditional networks when dealing with label noise in text classification.
Thus, we choose the representative BERT for further research and verify the generalization of SelfMix across different PLMs in Section~\ref{sec:5}.

\subsection{Baselines}
We compare SelfMix with the following baselines:
(1)~BERT, which trains the model with the cross-entropy loss without any denoising strategy;
(2)~Co-Teaching~\citep{coteaching}, which trains two models simultaneously and lets each model sample small-loss instances to teach the other model for further training;
(3)~Co-Teaching+~\citep{coteaching+}, which updates on disagreement data on the basis of the original Co-teaching;
(4)~SCE~\citep{SCE}, which boosts Cross Entropy symmetrically with Reverse Cross Entropy (RCE) for robust learning;
(5)~ELR~\citep{ELR}, which designs a regularization term to prevent memorization of the false labels; 
(6)~Confident-Learning~\citep{confidentlearning}, which estimates noise distribution by cross-validation and then trains a new model on clean data;
(7)~NM-Net~\citep{CIKM21} is one of the few representative works which jointly train a classifier and a noise model using a denoising loss;
(8)~$\text{CORES}^{2*}$~\citep{CORE} is a method for instance-dependent label noise, which progressively sieves out corrupted examples with a confidence regularization and applies semi-supervised learning for consistency training.
We implement them based on the standard BERT Encoder~\citep{bert} with reference to their public code and make comparisons under the same setting.

\subsection{Implementation Details}
There are three hyper-parameters to tune in SelfMix (the hyper-parameters of BERT are set as default and remain unchanged), the threshold $\tau$ for GMM to divide the data, and the weights $\lambda_p,\lambda_r$ for two special loss functions. 
We choose 0.5 as the threshold $\tau$ and keep it the same under different settings.
$(\lambda_p,\lambda_r)$ is demonstrated right besides the name of datasets in Table~\ref{table:asym}-\ref{table:idn}.
The performance can be more satisfactory if we specify the~$(\lambda_p,\lambda_r)$ for each setting. 
However, it is unfair to the methods that use few hyper-parameters, so we try to keep them the same.
Other settings like learning rate ($10^{-5}$), optimizer (Adam), and batch size (32) keep the same for all the methods and tasks.
For SelfMix, we warm up the model for 2 epochs under asymmetric noise and 5000 samples under instance-dependent noise.
Considering that the training data is noisy, we report the test accuracy of the best and last epochs over all 6 epochs rather than setting a clean validation set. And this is a commonly used metric in other related works.
All the results are the average of five runs.
Our noise generation code and more details can be found in our public code.

\begin{table*}[ht]\small
\centering
\renewcommand{\arraystretch}{1.2}
  \renewcommand\tabcolsep{7pt}
  \begin{tabular}{lc|cccc|cccl}
    \toprule
    Dataset~/~$(\lambda_p,\lambda_r)$&&\multicolumn{4}{c|}{AG-News~$(0.0,0.3)$}&\multicolumn{4}{c}{IMDB~$(0.0,0.3)$}\\
    \midrule
    Noise Ratio (\%)& &10 &20 &30 &40 &10 &20 &30 &40  \\
    \midrule
    \multirow{2}*{BERT} 
            & best &88.24&83.67&77.61&72.73 & 90.44 &83.07&79.52&76.59 \\ 
            & last &87.76&82.28&74.80&69.04 & 90.43 &80.26&70.43&60.59 \\
    \hline
    \multirow{2}*{BERT+Co-Teaching} 
            & best &88.62&84.63&78.40&73.14&90.00&83.64&79.70&76.09 \\
            & last &87.74&83.87&77.01&70.58&89.71&83.01&76.50&69.07 \\
    \hline
    \multirow{2}*{BERT+Co-Teaching+} 
            & best & \textbf{88.72} & 84.62 & 80.75 & 78.94 & 89.92 & \textbf{85.65} & 82.72 & 80.23 \\
            & last & \textbf{88.33} & 83.64 & 77.70 & 74.72 & 89.20 & \textbf{84.45} & 79.13 & 75.20 \\
    \hline
    \multirow{2}*{BERT+SCE} 
            & best &88.43&84.09&78.81&73.11 & 90.23& 84.30&80.60 & 75.76 \\
            & last &87.86&83.55&76.49&69.09 & 90.04& 82.59&75.46 & 67.94 \\
    \hline
    \multirow{2}*{BERT+ELR} 
            & best &88.45&83.41&77.77&72.97 & \textbf{90.60} &83.44&79.29& 76.10 \\
            & last &88.05&82.25&75.26&69.12 & \textbf{90.44} &80.91&71.81& 63.04 \\
    \hline
    \multirow{2}*{BERT+Confident-Learning} 
            & best &88.52&83.70&77.49 &71.58 & 90.09& 83.45&79.34 & 74.14 \\
            & last &88.20&83.23&75.97 &70.62 & 89.98& 82.12&75.76 & 69.05 \\
    \hline
    \multirow{2}*{BERT+NM-Net} 
            & best & 88.25 &83.19&76.60&72.31 & 90.05&83.28&79.54&75.85 \\
            & last & 87.92 &82.89&75.49&69.91 & 89.83&81.79&74.44&69.37 \\     
    \hline
    \multirow{2}*{BERT+$\text{CORES}^{2*}$} 
            & best &87.98&84.45&81.12&78.20 & 89.99&83.35&79.62&76.20 \\
            & last &86.76&82.79&78.67&75.39 & 73.39&62.90&55.47&58.16 \\
    \hline
    \multirow{2}*{BERT+SelfMix} 
    & best & 88.45 & \underline{\textbf{86.82}} & \underline{\textbf{86.72}} & \underline{\textbf{83.99}} & 90.31 & \textbf{85.49} & \underline{\textbf{84.38}} & \underline{\textbf{82.76}} \\
    & last &87.64 & \textbf{85.96} & \underline{\textbf{86.38}} & \underline{\textbf{83.67}} & 86.70 & \textbf{84.14} & \underline{\textbf{83.18}} & \underline{\textbf{78.94}} \\
  \bottomrule
\end{tabular}
  \caption{Average test accuracy (\%) of five runs on the AG-News and IMDB datasets under different ratios of instance-dependent noise. The results with outstanding improvement over the base model are bolded, and underline values indicate the statistically significantly better (by paired bootstrap test, $p < 0.05$) performances across the board.}
\label{table:idn}
\end{table*}

\begin{table}\small
  \centering
\renewcommand{\arraystretch}{1.2}
  \renewcommand\tabcolsep{3pt}
  \label{tab:freq}
  \begin{tabular}{lc|ccc}
    \toprule
    Dataset & &Trec&AG-News&IMDB \\
    \midrule
    \multirow{2}*{SelfMix w/o $\mathcal{L}_\mathcal{P}$} & best & 89.40  & 87.57 & 89.55\\
                                 & last & 85.04  & 83.21 & 87.40\\     
    \cline{1-5}
    \multirow{2}*{SelfMix w/o $\mathcal{L}_\mathcal{R}$} & best & 91.56  & 89.66 & 85.54\\
                                 & last & 88.28  & 87.73 & 75.98\\
    \cline{1-5}
    \multirow{2}*{SelfMix w/o mixup} & best & 91.52  & 89.51 & 88.23\\
                                   & last & 87.04  & 84.82 & 86.17\\               
    \cline{1-5}
    \multirow{2}*{SelfMix} & best & \textbf{94.12} & \textbf{92.79} & \textbf{90.19}\\
                                     & last & \textbf{93.80} & \textbf{92.40} & \textbf{90.12}\\                                 
  \bottomrule
\end{tabular}
  \caption{Ablation study results (\%) on Trec, AG-News and IMDB under 40\% asymmetric label noise.}
\label{table:ablation}
\end{table}

\subsection{Main Results} 
\textbf{Asymmetric Noise.}
The effect of asymmetric noise is relatively small when data is sufficient due to the excellent performance of PLMs.
So we cut the datasets into a small size of 5000 for more precise comparison of the models' performance.
Table~\ref{table:asym} shows the results on three datasets under asymmetric noise.
$\text{CORES}^{2*}$ is designed to handle IDN, so we show its performance in Table~\ref{table:idn}.
Our proposed SelfMix outperforms the strong baselines in almost every setting.
Most models' performance drops steeply under a high noise ratio and data-insufficiency setting. 
However, SelfMix still holds a remarkable performance over this challenging scenario.
SelfMix does not achieve the best result under 20\% label noise on AG-News, but it is excusable since the base model already holds a good performance and there is not much difference between SelfMix and the best result.

\begin{figure}[t]
	\centering
	\subfigure[20\% idn on AG-News]{
		\begin{minipage}[t]{0.5\linewidth}
			\centering
			\includegraphics[width=1.5in]{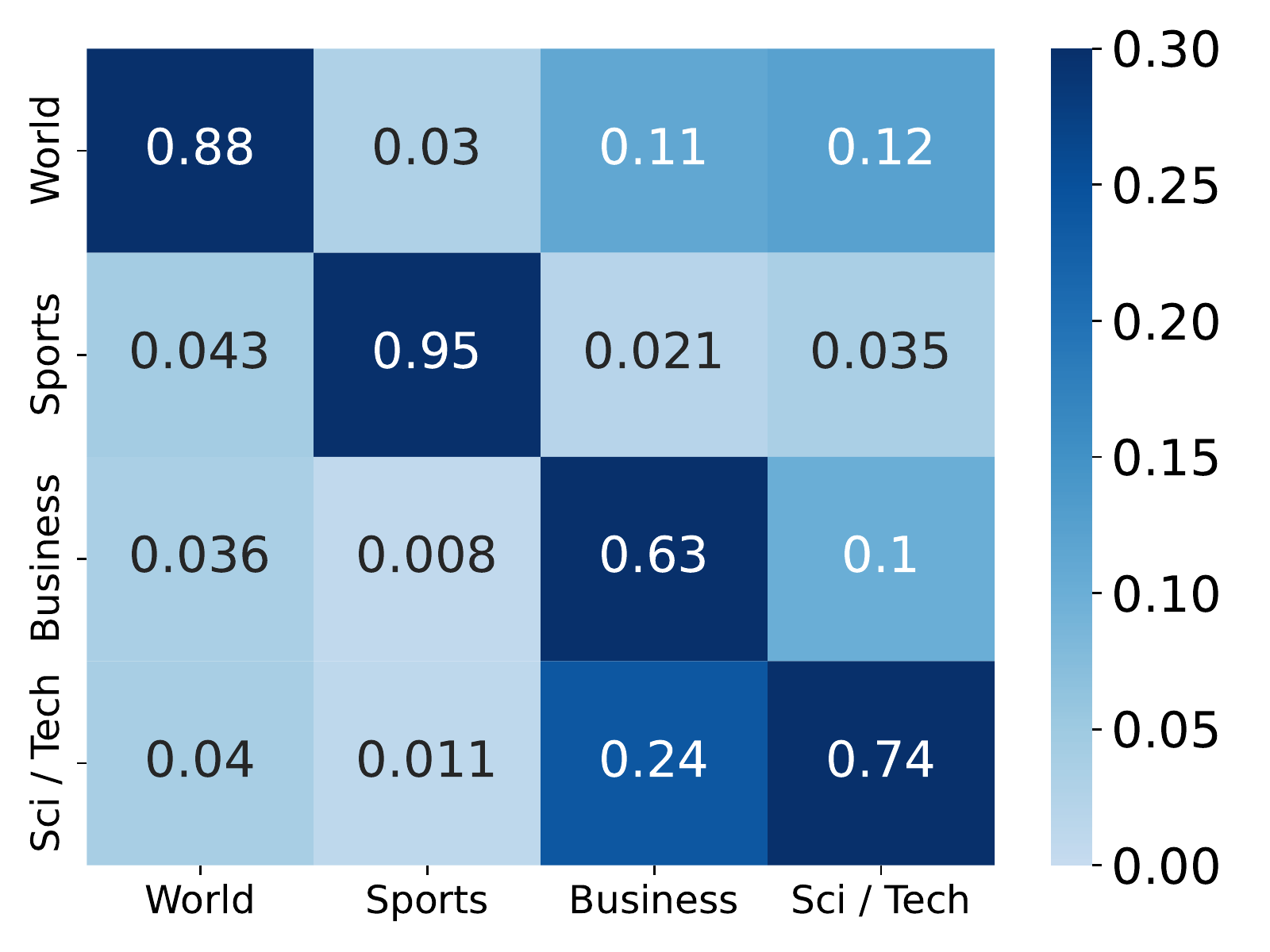}
		\end{minipage}
		\label{fig:2a}
	}%
	\subfigure[40\% idn on AG-News]{
		\begin{minipage}[t]{0.5\linewidth}
			\centering
			\includegraphics[width=1.5in]{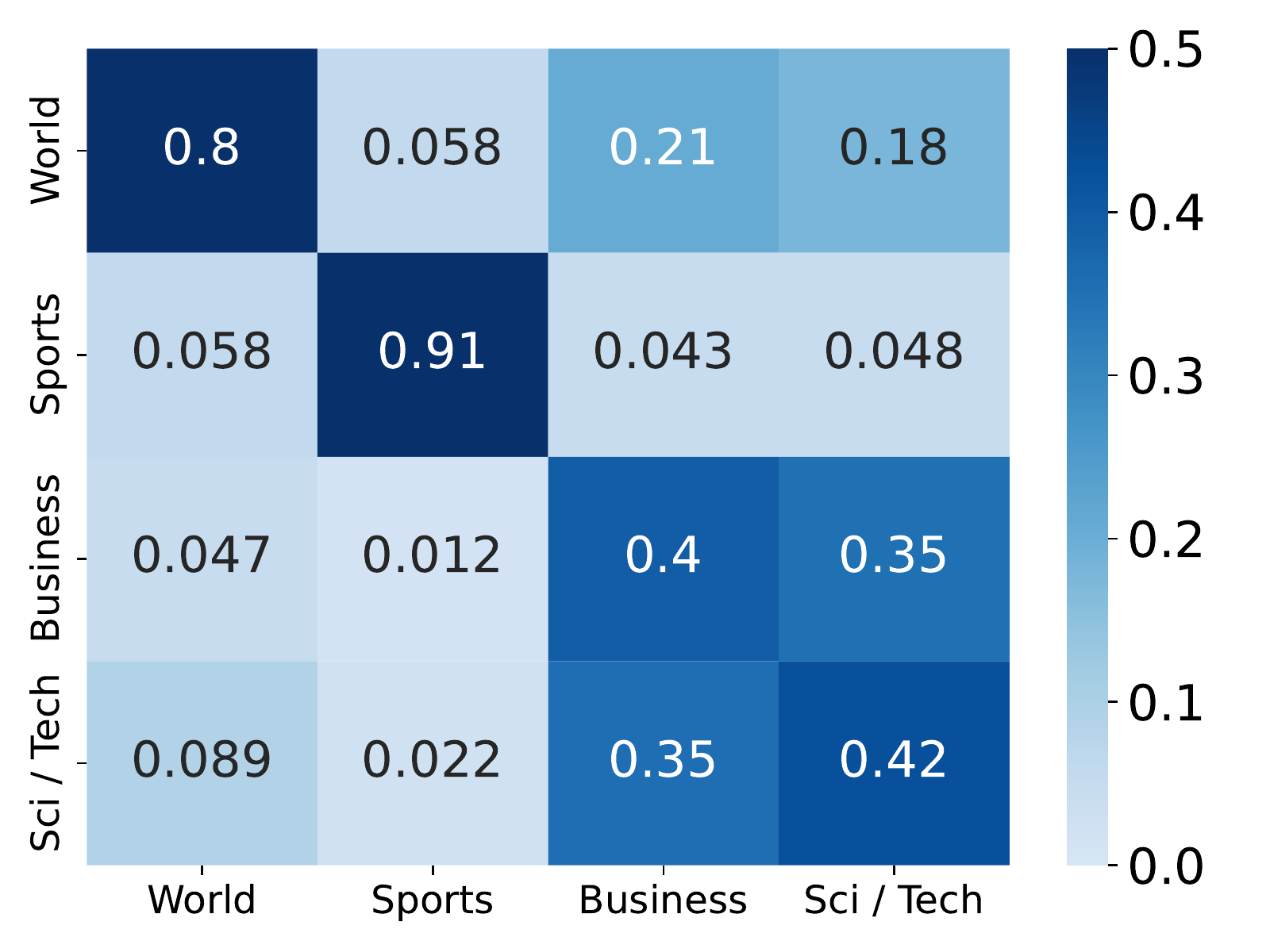}
		\end{minipage}
		\label{fig:2b}
	}%
	\centering
	\caption{The generated instance-dependent label noise distribution on AG-News, where the abscissa is the true label, and the ordinate is the observed label. }
	\label{fig:idn}
\end{figure}

\textbf{Instance-dependent Noise.}
~IDN is more close to real-world noise.
Following ~\citet{noise-effect}, we train an LSTM classifier on a small set of the original training data and flip the origin labels to the class with the highest prediction probability among other classes.
Trec dataset has only 5452 training samples and is extremely class-imbalance. So the number of clean samples may even be less than generated noisy samples in the long-tailed class under a high noise ratio, which makes the classification no sense.
Therefore, we only do experiments on IMDB and AG-News, and Figure~\ref{fig:idn} shows noise transition on AG-News.
Table~\ref{table:idn} presents the experimental results on IDN.
Some of the methods do not work properly since they were not designed for IDN and did not consider the discrepancy of loss distributions between different classes.
However, our proposed class-regularization loss can still make the samples distinguishable and SelfMix outperforms the strong baselines in most circumstances.

\begin{figure*}[ht]
    \centering
	\subfigure[Epoch 2 (Warmup/Base)]{
		\begin{minipage}[t]{0.25\linewidth}
		\hspace{-0.5cm}
			\centering
			\includegraphics[width=1.7in]{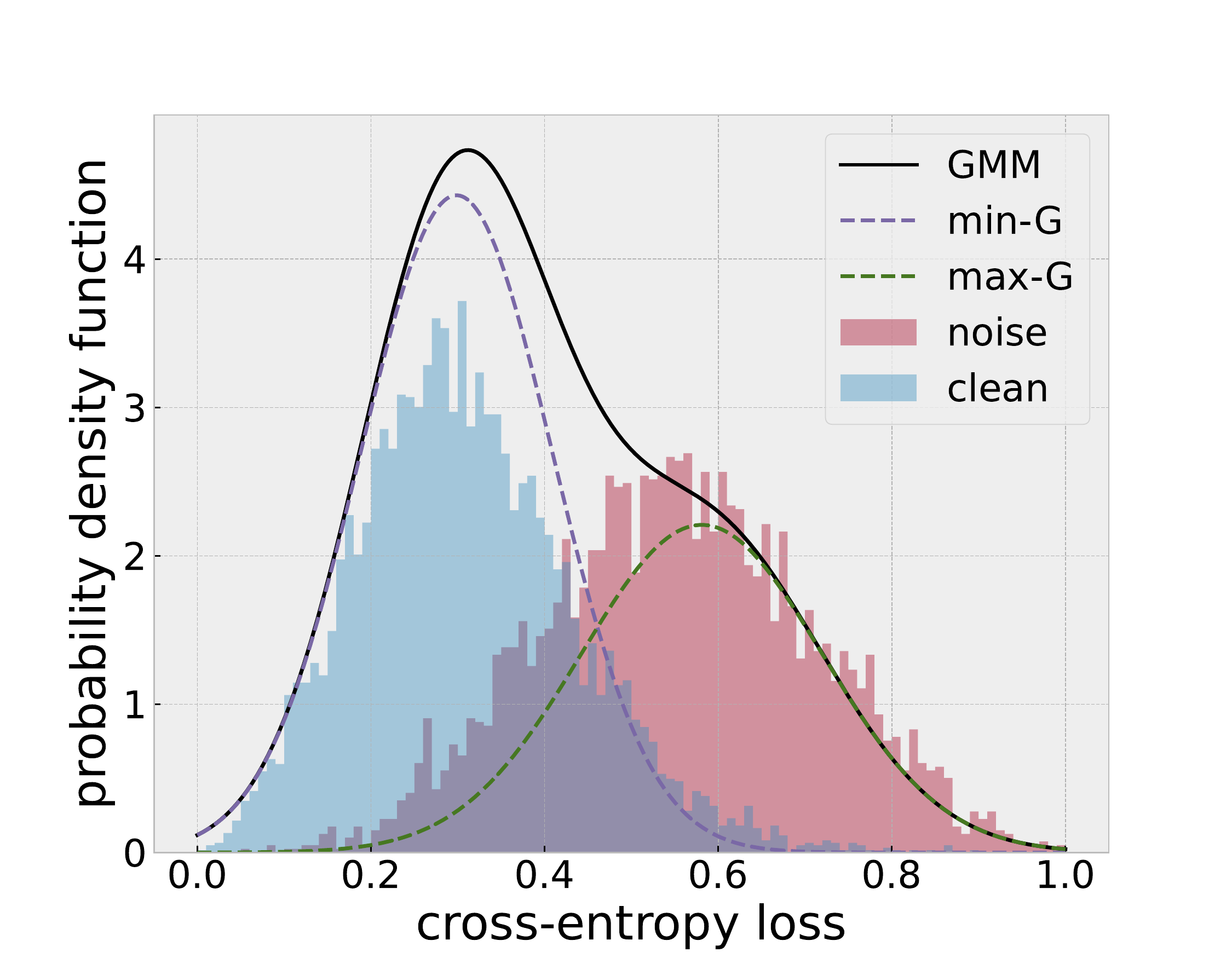}
		\end{minipage} 
		\label{fig:a}
	}%
	\subfigure[Epoch 6 (SelfMix)]{
	    \begin{minipage}[t]{0.25\linewidth}
	    \hspace{-0.5cm}
			\centering
			\includegraphics[width=1.7in]{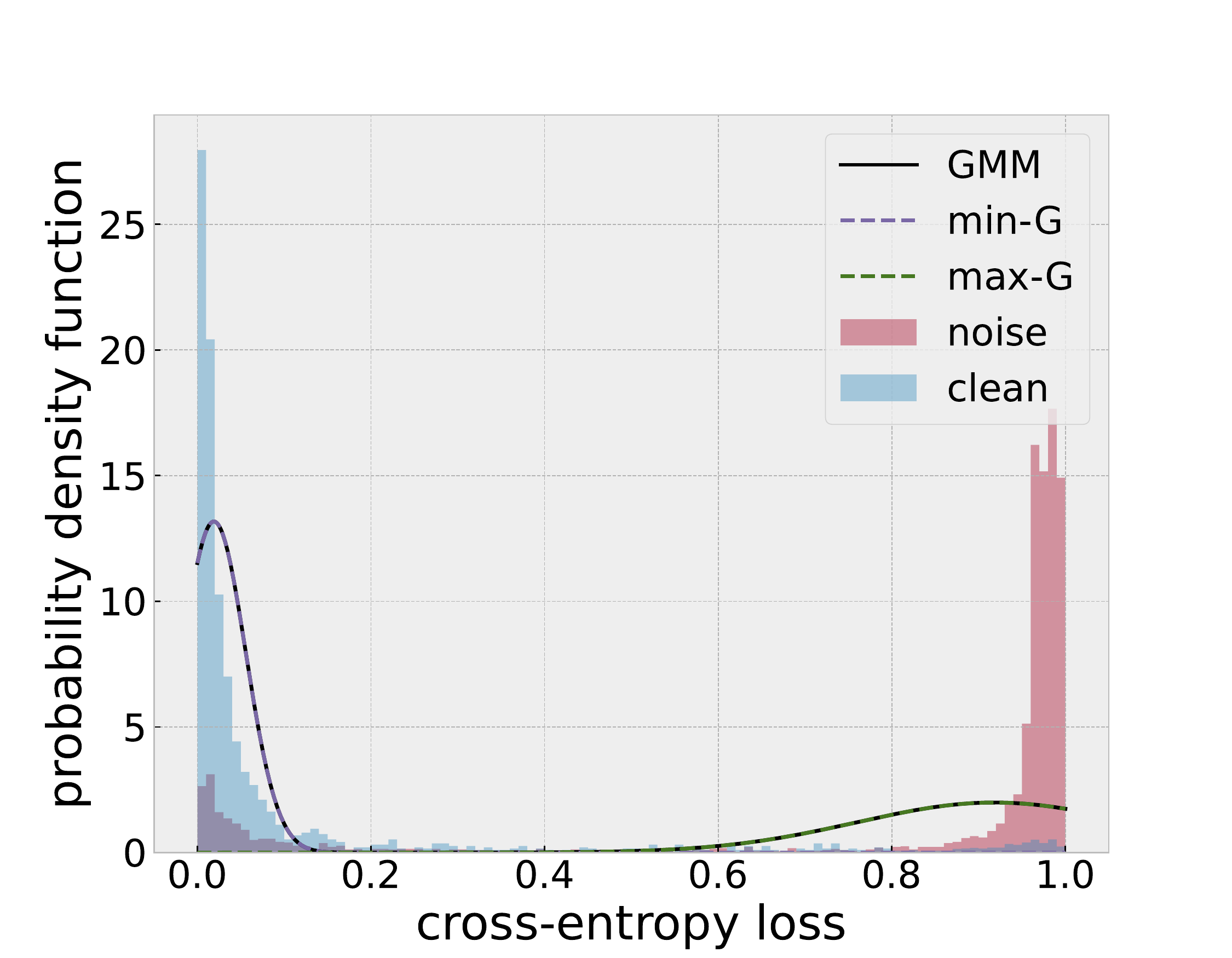}
		\end{minipage}
		\label{fig:b}
	}%
	\subfigure[Epoch 6 (Base)]{
		\begin{minipage}[t]{0.25\linewidth}
		\hspace{-0.5cm}
			\centering
			\includegraphics[width=1.7in]{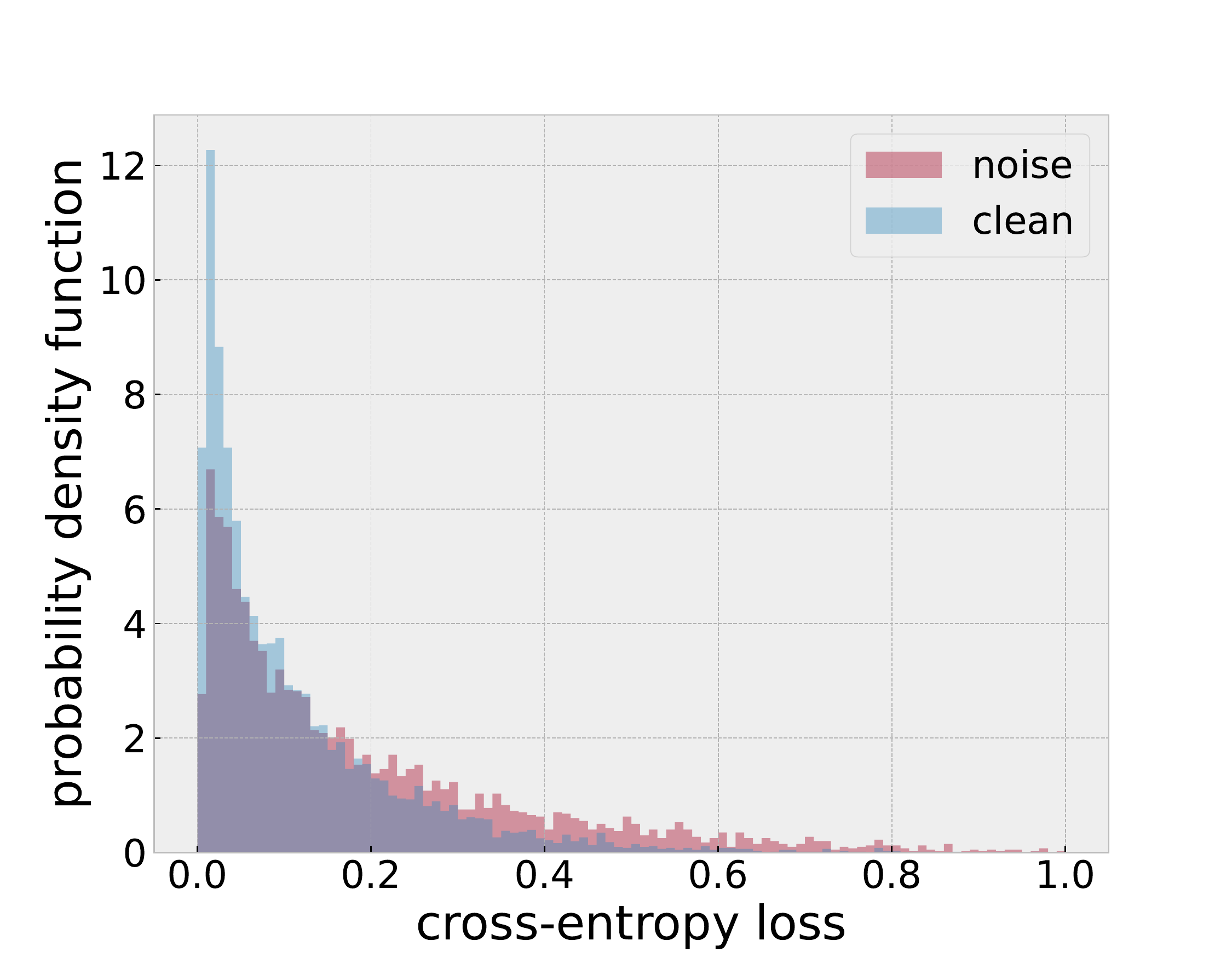}
		\end{minipage}
		\label{fig:c}
	}%
	\subfigure[Base-SelfMix]{
		\begin{minipage}[t]{0.25\linewidth}
		\hspace{-0.5cm}
			\centering
			\includegraphics[width=1.5in]{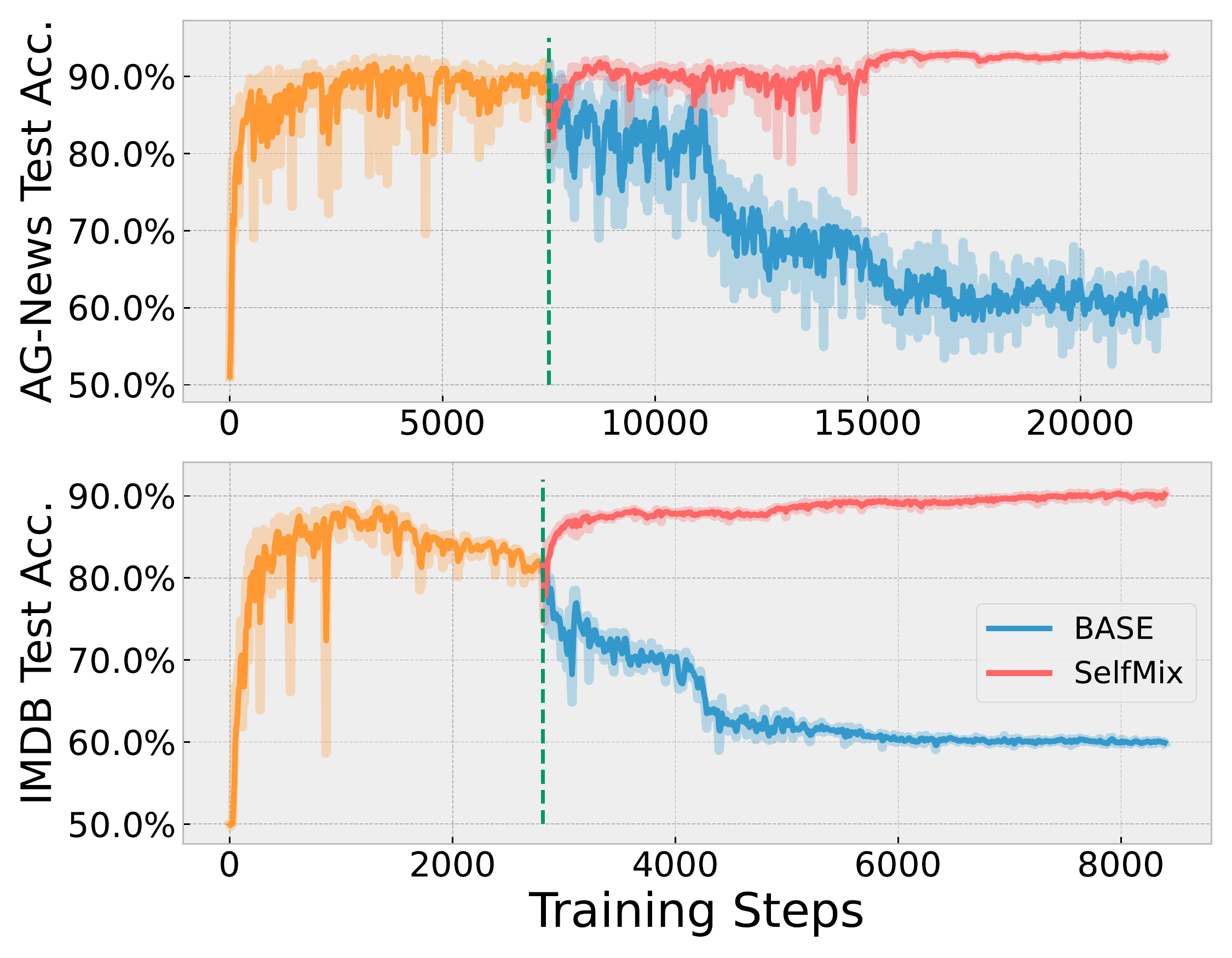}
		\end{minipage}
		\label{fig:d}
	}%
	
    \subfigure[Epoch 1 (Warmup/Base)]{
		\begin{minipage}[t]{0.25\linewidth}
		\hspace{-0.5cm}
			\centering
			\includegraphics[width=1.7in]{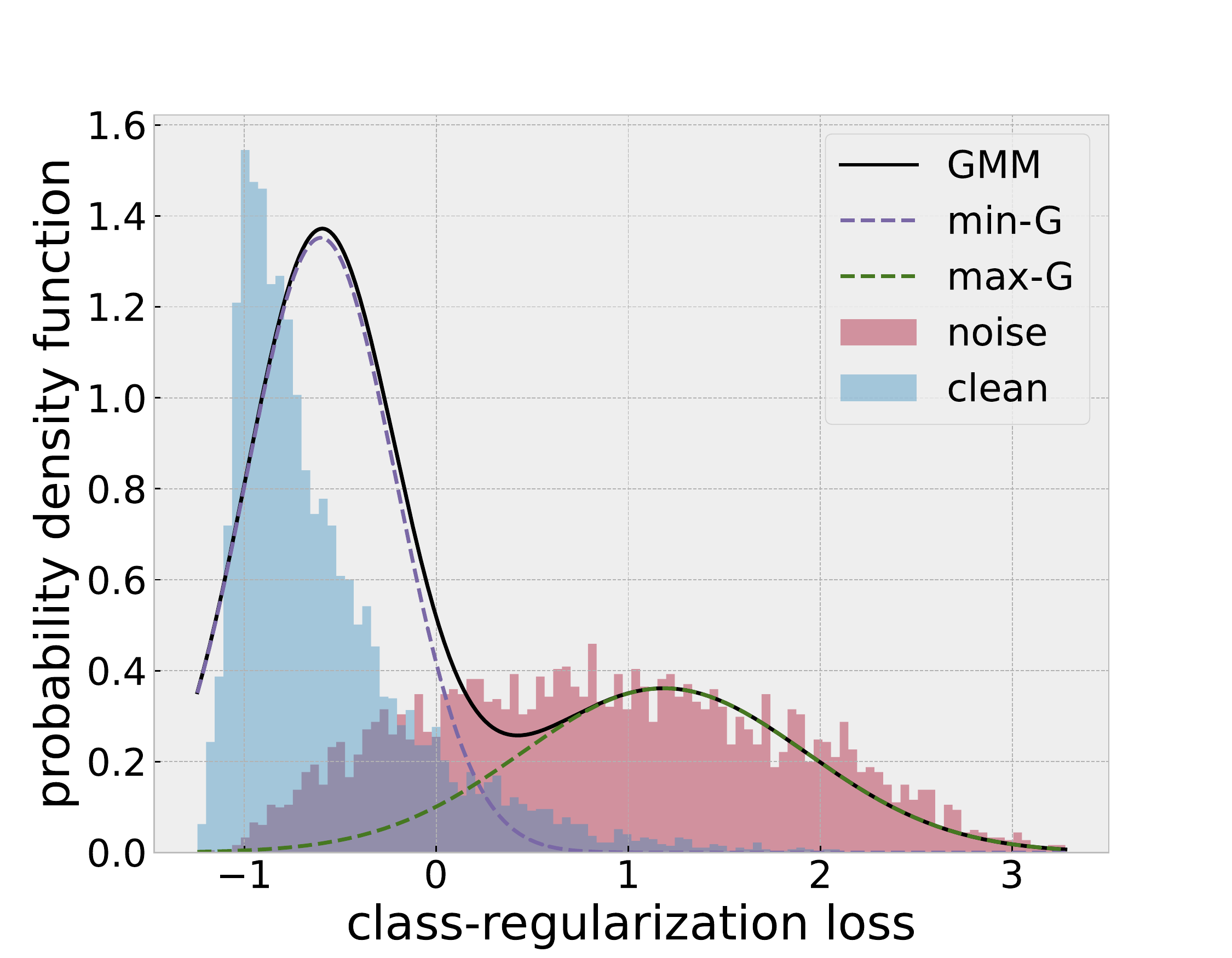}
		\end{minipage}
		\label{fig:e}
	}%
	\subfigure[Epoch 6 (SelfMix)]{
		\begin{minipage}[t]{0.25\linewidth}
		\hspace{-0.5cm}
			\centering
			\includegraphics[width=1.7in]{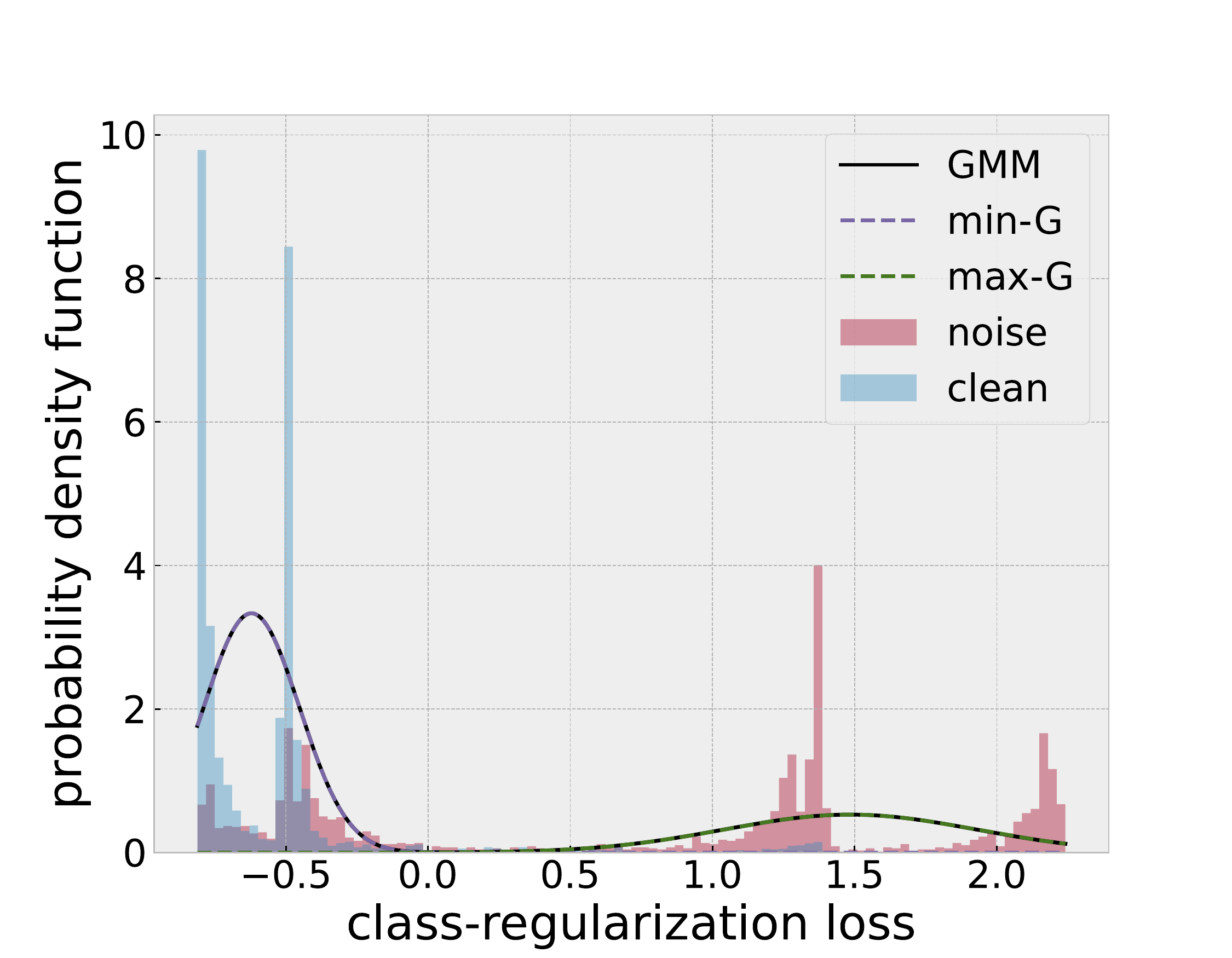}
		\end{minipage}
		\label{fig:f}
	}%
	\subfigure[Epoch 6 (Base)]{
		\begin{minipage}[t]{0.25\linewidth}
		\hspace{-0.5cm}
			\centering
			\includegraphics[width=1.7in]{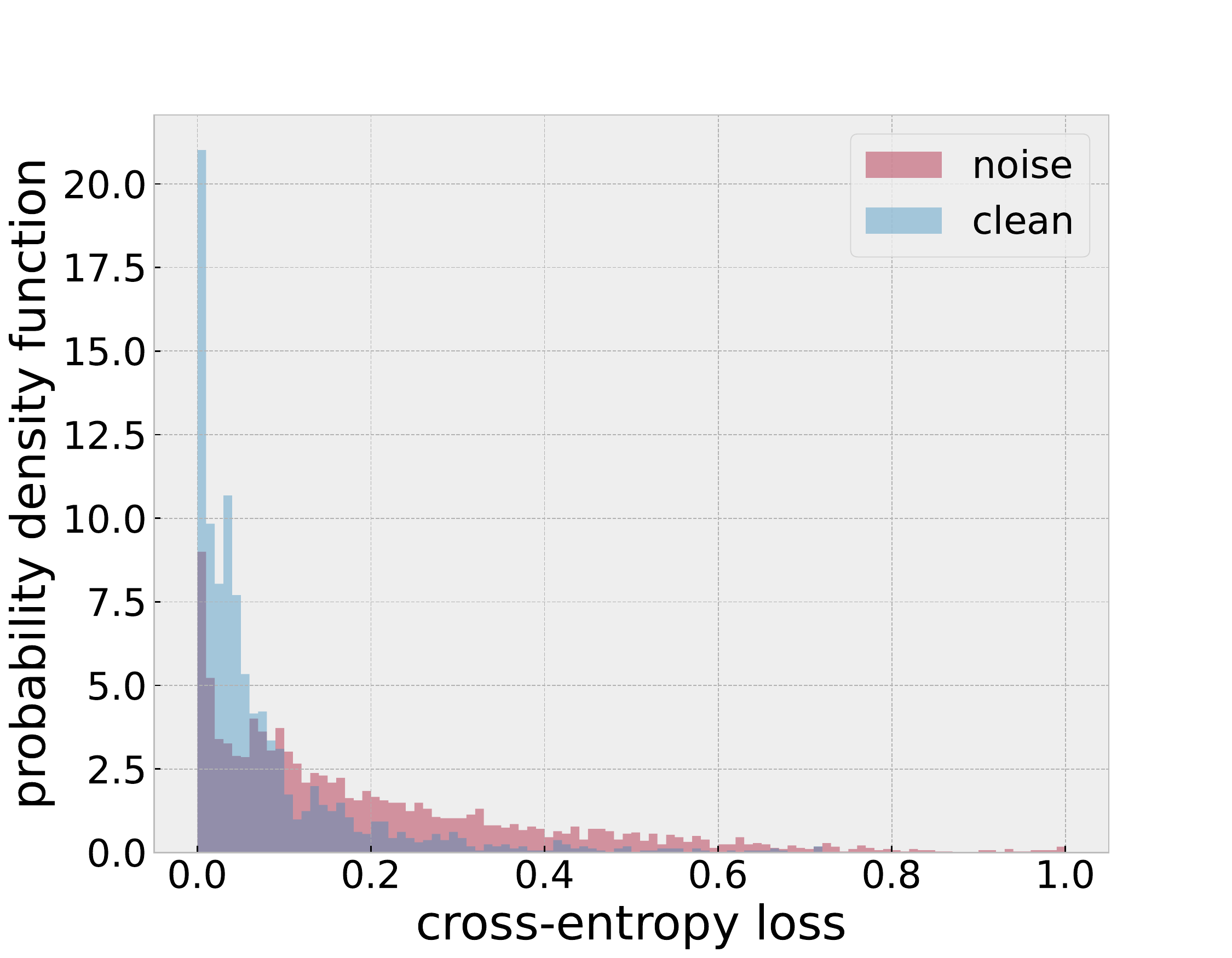}
		\end{minipage}
		\label{fig:g}
	}%
	\subfigure[Base-SelfMix]{
		\begin{minipage}[t]{0.25\linewidth}
		\hspace{-0.5cm}
			\centering
			\includegraphics[width=1.5in]{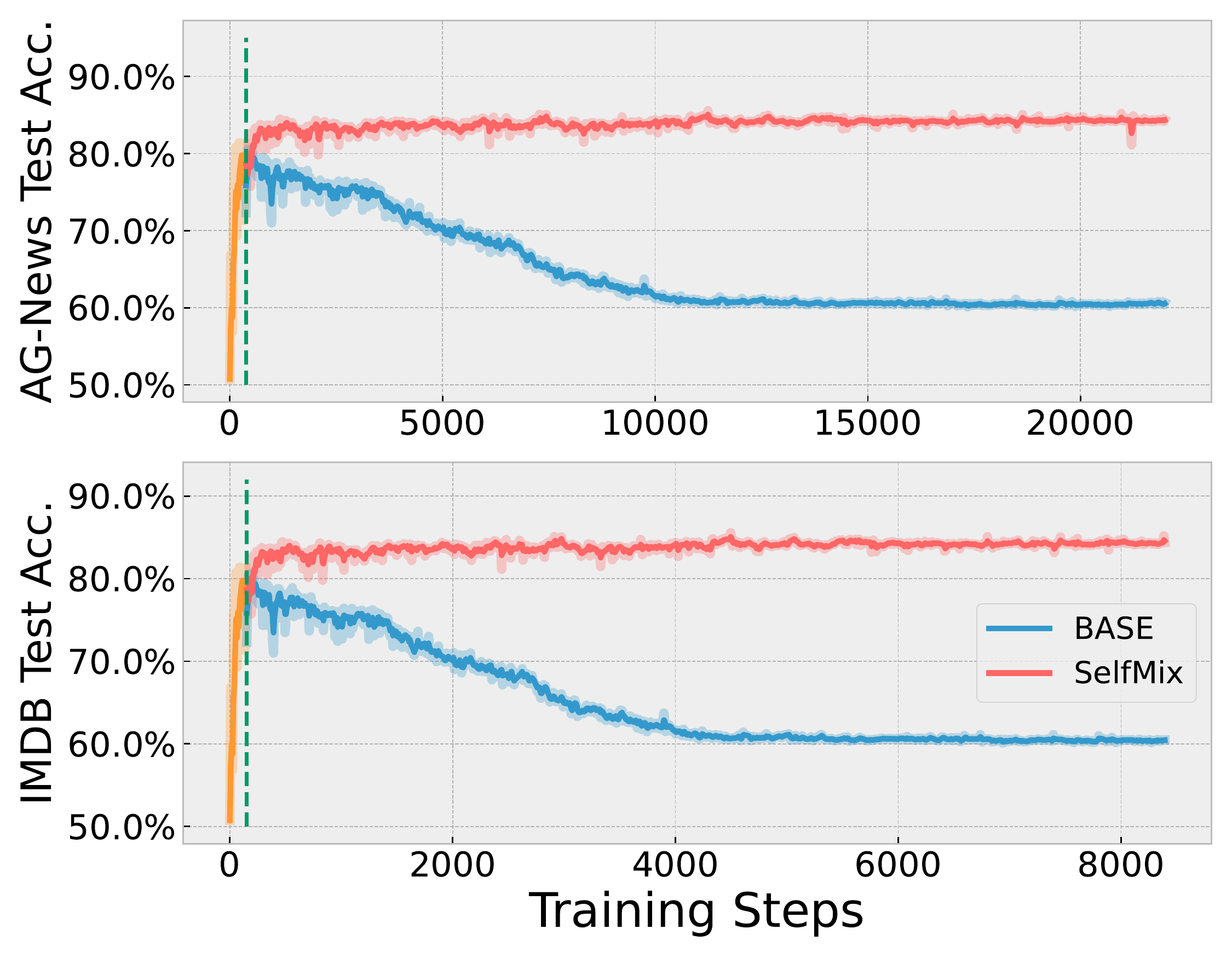}
		\end{minipage}
		\label{fig:h}
	}%
	\centering
	\caption{ (a-c) the loss distributions of SelfMix/Base on IMDB under 40\% asymmetric noise in different stages; (d) the test accuracy of every few training steps under 40\% asymmetric noise; (e-g) the loss distributions of SelfMix/Base on IMDB under 40\% instance-dependent noise in different stages; (h) the test accuracy of every few training steps under 40\% instance-dependent noise.
	}
	\label{fig:analysis}
\end{figure*}

\section{Analysis and Discussion}
\label{sec:5}

To make a more comprehensive analysis of our proposed strategies, we offer fine-grained experiments and visualization to answer the following research questions (RQs):
(1) Can GMM actually distinguish the noisy samples on textual data?
(2) How well can SeflMix help prevent the model from overfitting the noisy labels?
(3) SelfMix utilizes more than one component. Does each of them contribute to the final performance?
(4) Can SelfMix be applied to other pre-trained models except BERT?
(5) Noise and outliers both might have higher loss in early stages. While examples with noisy labels are useless or detrimental while training, how do we make sure with GMMs we don't filter out outliers in this approach?

\textbf{Answer 1:} 
We demonstrate the loss distributions of the clean samples and noisy samples on IMDB under 40\% asymmetric noise in Figure~\ref{fig:analysis} (a-c) and IDN in Figure~\ref{fig:analysis} (e-g).
Consistent with \citet{ELR}, the model tends to learn clean data during an early learning phase, and the 2-component GMM almost perfectly fits the loss distribution to distinguish the clean and noisy samples.
During training, the loss output by SelfMix is getting more polarized while the base model has already overfitted the wrong labels.
Notably, the cross-entropy loss values of different classes vary greatly under instance-dependent noise. 
And our proposed class-regularization loss can help GMM better isolate these distributions in each class. 

\textbf{Answer 2:}
We record the test accuracy for every few mini-batches and show the learning process on AG-News (120k samples) and IMDB (45k samples) under 40\% asymmetric/instance-dependent noise in Figure~\ref{fig:d}/\ref{fig:h}. 
The left side of the green vertical dotted line records the warm-up stage of SelfMix, which is the same as the base model.
From the right side, we can observe that the base model overfits the noisy samples quickly. At the same time, SelfMix can keep learning and performs better, which may benefit from the effective sample selection and mixup training.
The loss distributions in Figure~\ref{fig:analysis} can also prove that.
We have an interesting observation: The training process under IDN is more stable than asymmetric noise.
We assume that the randomness in asymmetric noise breaks the stability of the map from features to output probability. 
While for IDN, there still exists a learnable map from input features to output labels, which makes the learning process no different from a standard text classification from another perspective.

\textbf{Answer 3:}
We remove each sub-method used in SelfMix respectively and check the test accuracy to see whether each component of our proposed method contributes to the task (Table~\ref{table:ablation}). 
We observe that each component can significantly contribute to the final performance.
$\mathcal{L}_{\mathcal{P}}$ and mixup training play a more critical role against overfitting since the results of the last epoch fall sharply without these two mechanisms.
Another unexpected but reasonable observation is the precipitous dropping result without $\mathcal{L}_\mathcal{R}$ on IMDB under 40\% noise.
SelfMix utilizes dropout as an alternative to prevent confirmation bias in self-training.
However, 40\% asymmetric label noise blurs the class boundary.
It inevitably leads to the inconsistency between implicit sub-models, which is more pronounced on the binary classification dataset IMDB, and $\mathcal{L}_\mathcal{R}$ just constraints the divergence between sub-models.

\textbf{Answer 4:}
To verify the effectiveness of our proposed SelfMix on other PLMs, we perform experiments on RoBERTa.
Table~\ref{table:roberta} shows the significant improvement brought by SelfMix.

\textbf{Answer 5:}
1).Outlier refers to a data point that is significantly dissimilar to other data points or a point that does not imitate the expected typical behavior of the other points \citep{1wang2019progress}, which has some similarities with the concept of noisy samples. Most noisy sample filtration methods are constructed based on the consumption or phenomenon that noisy samples behave differently from other data points during training. With the overlapped concept and the similar consumption/phenomenon in distinguishing noisy samples and outliers from other data points, many outlier detection methods resemble the noisy filtration ones \citep{2wu2020topological, 3knox1998algorithms}, i.e., they view a point as an outlier/noisy sample if it is far away from its nearby neighbors in the representation space. As one of the most representative strategies in both noisy sample filtration and outlier detection, the conventional GMMs used in this paper is difficult to distinguish precisely the outliers and noisy samples.
2).Actually, excluding outliers along with the filtration of noisy samples from clean data may not be harmful. As mentioned by \citet{4zhu2008active}, these selected outliers (i.e., unlabeled examples) have high uncertainty and cannot provide much help to learners. \citet{5shin2006enhanced} also show that excluding outliers from the noisy training data significantly improves the performance of the centroid-based classifier. Moreover, \citet{6carlini2019distribution} have made a comprehensive study on what impact outliers exactly bring to deep neural networks. For the tasks of image classification, outliers/hard samples are only helpful when training on easy-to-learn data. In this paper, the mixed data is challenging enough that it may not benefit from keeping these filtrated outliers in training. 
From another perspective, outliers in textual data appear to be inherently misleading or ambiguous examples located on the clustering boundary. The mixup strategy of this work can generate adequate samples around the boundary.

\begin{table}\small
  \centering
\renewcommand{\arraystretch}{1.2}
  \renewcommand\tabcolsep{2.5pt}
  \begin{tabular}{lc|c|cc|cc}
    \toprule
    Dataset&& Trec &\multicolumn{2}{c|}{AG-News}&\multicolumn{2}{c}{IMDB}\\
    \midrule
    Noise Type & & Asym & Asym & IDN & Asym & IDN \\
    \midrule
    \multirow{2}*{RoBERTa} & best & 92.12 & 92.74& 72.49 & 90.54&74.09\\
                        & last   & 86.56  & 89.43& 69.94  & 80.60&60.50\\
    \hline
    \multirow{2}*{RoBERTa+Ours}  & best & \textbf{94.88} & \textbf{92.81} & \textbf{84.44} & \textbf{92.33} & \textbf{91.19}\\
                            & last & \textbf{94.64} & \textbf{92.15} & \textbf{82.87} & \textbf{92.14} & \textbf{91.10}\\               
  \bottomrule
\end{tabular}
  \caption{Test performances (\%) on RoBERTa under 40\% asymmetric/instance-dependent noise.}
\label{table:roberta}
\end{table}

\section{Conclusions}
This paper presents SelfMix to handle label noise on textual data. It uses the Gaussian mixture model for sample selection and applies EmbMix for semi-supervised learning.
Unlike the mutual distillation methods requiring co-training or model assembling, the proposed framework needs only a single model with dropout mechanism and utilizes two specific regularizations.
Extensive experiments conducted on three representative text classification datasets under different noise settings indicate that SelfMix achieves a significant improvement over strong baselines. However, the proposed framework does not explicitly distinguish outliers and label noise. The future work includes exploring the different roles the noisy data and outliers play and applying our method to other supervised natural language tasks like Named Entity Recognition.

\section*{Acknowledgments}
We thank the anonymous reviewers for their constructive comments. The first three authors (Dan Qiao, Chenchen Dai, Yuyang Ding) contribute equally to this work. 
Juntao Li is the corresponding author.
This work was supported by Alibaba Innovative Research (AIR) Grant and Project Funded by the Priority Academic Program Development of Jiangsu Higher Education Institutions.

\bibliography{anthology,custom}
\bibliographystyle{acl_natbib}

\end{document}